\documentclass{article} 
\usepackage{iclr2024_conference,times}

\usepackage{amsmath,amsfonts,bm}

\def\eqref#1{equation~\ref{#1}}

\def\1{\bm{1}}

\DeclareMathAlphabet{\mathsfit}{\encodingdefault}{\sfdefault}{m}{sl}
\SetMathAlphabet{\mathsfit}{bold}{\encodingdefault}{\sfdefault}{bx}{n}

\usepackage{hyperref}
\usepackage{url}
\usepackage{xcolor,colortbl}
\usepackage{pifont}
\definecolor{row-highlight-color}{HTML}{e8eef4}
\usepackage{textcomp, gensymb}
\usepackage{algorithm}
\usepackage{algpseudocode}
\usepackage{adjustbox}
\usepackage{booktabs}

\title{A Concise Tiling Strategy for Preserving \\ Spatial Context in Earth Observation Imagery}

\author{Ellianna Abrahams, Tasha Snow, Matthew R. \\
Department of Statistics\\
University of California, Berkeley\\
Berkeley, CA 94720, USA \\
\texttt{\{ellianna, fernando.perez\}@berkeley.edu} \\
\And
\hspace{-50pt}
Siegfried, \& Fernando P\'{e}rez\\
Department of Geophysics \\
Colorado School of Mines \\
Golden, CO 80401, USA\\
\texttt{\{tsnow,siegfried\}@mines.edu} 
}

\iclrfinalcopy 
\begin{document}

\maketitle

\begin{abstract}
We propose a new tiling strategy, \textbf{Flip-n-Slide}, which has been developed for specific use with large Earth observation satellite images when the location of objects-of-interest (OoI) is unknown and spatial context can be necessary for class disambiguation. Flip-n-Slide is a concise and minimalistic approach that allows OoI to be represented at multiple tile positions and orientations. This strategy introduces multiple views of spatio-contextual information, without introducing redundancies into the training set. By maintaining distinct transformation permutations for each tile overlap, we enhance the generalizability of the training set without misrepresenting the true data distribution. Our experiments validate the effectiveness of Flip-n-Slide in the task of semantic segmentation, a necessary data product in geophysical studies. We find that Flip-n-Slide outperforms the previous state-of-the-art augmentation routines for tiled data in all evaluation metrics. For underrepresented classes, Flip-n-Slide increases precision by as much as 15.8\%. 
\end{abstract}

\section{Introduction}

The volume of geospatial satellite imagery has grown rapidly in the past decade. Semantic segmentation presents a promising opportunity for rapidly parsing meaningful scientific understanding from these images. Despite the remarkable accomplishments of deep neural networks for such segmentation tasks \citep{ronneberger_2015a,chen_2019,tan_2020,amara_2022}, these methods can underperform on data that have noisy or underrepresented labels \citep{shin_2011,guo_2019} or when one set of data representations is used for a wider set of downstream tasks \citep{yang_2018}. These are common challenges in Earth observation imagery. To overcome these issues, data augmentation is a widely adopted technique for generalizing a model fit to make better predictions by expanding the size and distribution of training data through a set of transformations \citep{vandyk_2001,hestness_2017}. In recent years, much focus has been given to upstream augmentation methods that address overfitting through data mixing \citep{zhang_2017, yun_2019, hong_2021, dabouei_2021} or proxy-free augmentations \citep{cubuk_2019, cubuk_2020, reed_2021,li_2023}---strategic approaches that expand the training data, but also execute unrealistic data transformations. Furthermore, limited attention has been given to investigating upstream augmentation techniques in the realm of learning on tiled imagery, an approach often employed to parse large images into smaller tiles to overcome the intractable size of the overall image for the GPU memory \citep{pinckaers_2018, huang_2019}. 

\begin{figure*}
  \centering
  \includegraphics[width=\linewidth]{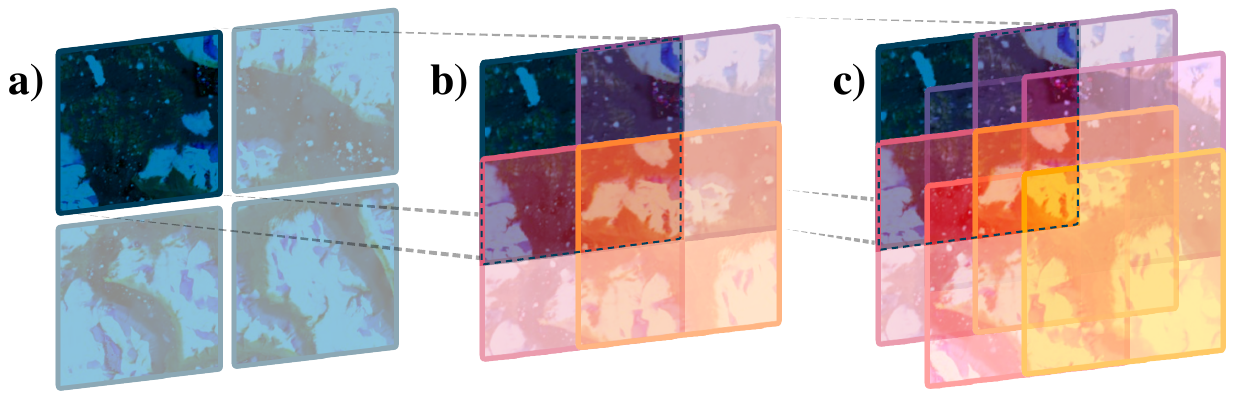}
  \caption{Flip-n-Slide's tile overlap strategy creates eight overlapping tiles for any image region more than a 75\% tile threshold away from the overall image edge. Three tiling strategies, shown in false color to illustrate overlap, are visualized here. a) Tiles do not overlap. b) The conventional tile overlap strategy, shown at the recommended 50\% overlap. c) Flip-n-Slide includes more tile overlaps, capturing more OoI tile position views for the training set.}
  \label{fig:overlap}
\end{figure*}

Tiling techniques in scientific applications require intentional augmentation choices, as certain transformations are unphysical and therefore not useful to learn. Spatial context is needed to disambiguate between classes with similar channel outputs and surface textures \citep{pereira_2021}. This is particularly true in scientific use-cases where tasks like classification are employed to answer wider questions about rare or relatively unknown phenomena. Although most machine learning methods demonstrate robust performance with well-represented phenomena, this performance degrades when training data fails to accurately capture poorly represented features or the structure of the problem. Human experts are better able to identify rare and unknown phenomena by relying on domain context, such as the semantic or temporal proximity of other classes \citep{wang_2023}; it stands to reason that machines could similarly use context for similar purposes.

To maintain access to context after tiling, the current state-of-the art employs tiling as a sliding window data augmentation, overlapping tiles by a significant percentage to extend dataset size and expose the model to multiple contextual views \citep{ronneberger_2015a,unel_2019,zeng_2019,akyon_2022} at training time, at test time, or during both. However, when applied upstream, this approach leads to redundancies within the training set, as many pixel windows are repeated more than once; for objects-of-interest (OoI) that are the size of the window slide, spatial context is limited beyond the singular sliding parameter \citep{zeng_2019,reina_2020}. This leads to the question: \textbf{How can we best tile and augment large images with limited physically realistic transformations without losing important spatio-contextual information in the tiling process?} Further: \textbf{Can we achieve this without creating the redundancies that occur in simply overlapping tiles?}

To answer this question, we propose a concise augmentation strategy, \textbf{Flip-n-Slide}, built for use with large Earth observation images where: 1) tiling is necessary; 2) data transformations must be limited to rotations and reflections to be realistic; and 3) there is no prior knowledge of the pixel locations for which spatial context will be necessary. We argue that physically realistic transformations of the data can be implemented \textit{alongside} the tiling overlap process, thereby removing redundancies when training convolutional neural networks (CNNs), in which orientation matters for learning \citep{ghosh_2018,szeliski_2022}. Our strategy allows us to create a larger set of samples without the superfluity of simply overlapping the tiles, enhancing downstream generalization. To achieve our goal, we slide through multiple overlap thresholds of the tiling window, exposing the model to more contextual views of each location (Figure \ref{fig:overlap}), and distinctly permute any overlapping windows to eliminate redundancy with other tiles that share pixels. We run segmentation experiments using our routine on the Land Cover of Canada (LCC) dataset \citep{latifovic_2020} to classify Landsat 8 satellite imagery \citep{eros_2013,roy_2014}. Our findings demonstrate that our strategy improves the segmentation performance of tiled data, especially for underrepresented classes, when compared to the conventional method (Figure \ref{fig:data_dist}c).

\section{Tiling and Augmentation Method}
We present the Flip-n-Slide algorithm, a data preprocessing strategy that tiles a large input image using a sliding window, providing eight sliding overlaps for every tile. Redundancies on overlap are eliminated using distinct, physically realistic transformations to permute each overlapping tile.

\subsection{Mathematical Formalism for the Flip-n-Slide Algorithm}

Flip-n-Slide follows a concise mathematical formulation to concurrently tile and transform a large image. In Algorithm \ref{alg:cap} we outline the formal process for implementing Flip-n-Slide on an input image, $I$. The algorithm is implemented in two stages. First a sliding window captures overlapping tiles, $t \in T$, at specified boundary ranges, $i\times j$, in the 2D image plane of $I$. The boundary ranges are given in $S$. This operation is specified in Line \ref{alg:one}. The tile boundaries are defined such that any given $x\times y$ pixel location sufficiently away from the edge of the overall image $I$ overlaps with eight tiles. The length of $i$ is equal to the length of $j$, creating a square. Each of these overlapping tiles is permuted correspondingly from a specified set of eight distinct transformations, $F$. This operation is specified in Line \ref{alg:two}. As augmentation is implemented concurrently with tiling, Flip-n-Slide must be implemented before the data streaming stage. This results in the constraint of CPU storage capacity required to accommodate the complete extended dataset. The overlap and transformation choices are explained in greater detail in Section \ref{subsec:hyperparam}. We release this algorithm as a versioned, documented, open-source \texttt{Python} package named \texttt{flipnslide}, available on \href{https://pypi.org/project/flipnslide/}{PyPI} and \href{https://github.com/elliesch/flipnslide}{GitHub}.

\subsection{Hyperparameter Selection for Tiling and Transformation}\label{subsec:hyperparam}
We justify the fixed parameter choices for Flip-n-Slide to meet the goal of preserving spatial context for OoI, while eliminating informational redundancies that have been introduced in past tiling strategies. Our approach has the added benefit of expanding the initial training distribution with physically realistic transformations, adding further downstream generalization to the model fit.  \\[-8px]

\noindent\textbf{Tile Size}: Similar to past methods \citep{ronneberger_2015a, unel_2019}, Flip-n-Slide employs a fixed tile size to enable ease-of-scale when processing petapixel datasets. The algorithmic method is independent of tile size as needs will vary depending on the size of OoI, allowing users to pick a size that is most appropriate for a specific use-case. However, we limit the algorithm to square tiles to make the data more efficient for use with CNNs. \\[-14px]

\noindent\textbf{Tile Overlap} \label{subsec:overlap}: Previous methods employ a single overlap threshold, generally recommended to be 50\%. Flip-n-Slide uses a strategy with multiple overlap thresholds, where a tile window slides along to capture a 0\%, 25\%, 50\%, and 75\% threshold on both spatial axes, leading to eight overlaps for any given $x \times y$ area that is more than a 75\% tile threshold away from the edge of the original large image. This method provides the model with multiple tile views of contextual information for OoI, leading to greater downstream tile-position-invariance. Additionally, this method reduces the overhead present in dynamic methods by removing the need for an upstream check that small objects are not sliced during augmentation; in Flip-n-Slide sliced objects that are smaller than the tile size will have unsliced representations in other overlapping tiles. \\[-14px]

\noindent\textbf{Transformation Permutations} \label{subsec:transform}: Rotations and reflections provide alternative views of spatial relationships to the convolutional kernel in CNNs \citep{vandyk_2001}. For each of the eight tile overlaps, Flip-n-Slide employs a distinct rotation and/or reflection transformation from the following set of permutations $\{0\degree, 90\degree, 180\degree, 270\degree, (0\degree, \rightarrow), (0\degree, \uparrow), (90\degree, \rightarrow), (90\degree, \uparrow) \}$, shown in Figure \ref{fig:permute}. This set avoids commutative and indistinct transformation compositions to reduce redundancies for tiles that overlap in some way on the same initial image position. Additionally, it is limited to physically realistic transformations of the original image bypassing any pixel-level effects that can be introduced when rotating square matrices to angles indivisible by $90\degree$.

\section{Experimental Setup}

We evaluate our strategy’s performance on a classification task with a benchmarked semantic segmentation approach and compare it with the current recommended tiling convention as a control. Across all experiments, we argue that our augmentation strategy exposes the model to enough semantic context to remove the need for augmentation and label averaging at training time. To test this hypothesis, we only tile the test image and do not add any further augmentations at test time.

\begin{figure}
  \centering
  \includegraphics[width=\linewidth]{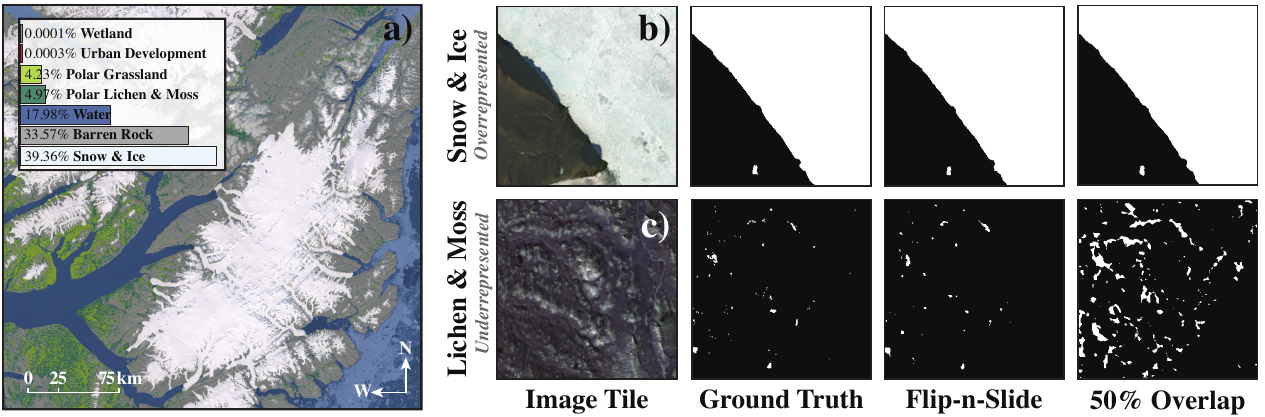}
  \caption{In Earth observation, classes can be extremely imbalanced as is shown here for Ellesmere Island, Nunavut, CA. a) Labels from the Land Cover of Canada dataset \citep{latifovic_2020} overlain on the corresponding Landsat satellite imagery \citep{eros_2013}. The legend shows the relative class distribution. Image tiles showing (b) an over-represented class (snow and ice) and (c) an under-represented class (lichen and moss) with the binary class masks from the ground truth data set, segmentation after training using the Flip-n-Slide tiling algorithm, and segmentation after training using the conventional tiling algorithm (50\% overlap). Although both algorithms perform well for the over-represented class case, Flip-n-Slide is more precise, by up to 15.8\%, than the conventional strategy (50\% tile overlap). 
  }
  \label{fig:data_dist}
\end{figure}

We perform our experiment on data from Ellesmere Island in the Canadian Arctic to ensure that our tiling strategy performs well on datasets in wilderness regions, which are an understudied focus of inquiry in established machine learning datasets, (Figure \ref{fig:data_dist}a). We use land cover classifications from the publicly available Land Cover of Canada (LCC) dataset \citep{latifovic_2020} and the corresponding images from NASA's Landsat 8 imagery \citep{eros_2013,roy_2014} since the LCC was derived from Landsat images. The data are classified by seven labeled classes (Snow \& Ice [39.36\% of the overall dataset], Barren Rock [33.57\%], Water [17.98\%], Polar Moss \& Lichen [4.97\%], Polar Grassland [4.23\%], Urban Development [0.0003\%], Wetland [0.0001\%]). There is high class imbalance in this data as is common in geophysical land cover segmentation problems, where often the OoI are among the least represented classes. 

We evaluate the performance of Flip-n-Slide on a semantic segmentation task, generating 12,800 256$\times$256 tiles from the dataset. All training experiments are conducted using the benchmarked UNet architecture \citep{ronneberger_2015a}, employed with the ADAM optimizer \citep{kingma_2017} at a learning rate of 0.001 and an unweighted cross-entropy loss. We train additional models on the two control tiling strategies to test the comparative performance, employing the same model architecture and optimization. The models undergo 300 training epochs with a batch size of 32.

\section{Experimental Results}\label{sec:results}
We follow the standard evaluation setup for semantic segmentation tasks using a UNet \citep{ronneberger_2015a}. We train a first model on tiles generated by employing the Flip-n-Slide strategy. As a comparative control, the second model is trained on non-overlapping tiles, and the third is trained on tiles generated using the conventional 50\% overlap strategy \citep{unel_2019, zeng_2019, reina_2020, akyon_2022}. We test all three trained models on three random draws of non-overlapping tiles from the same geographic location, and share the performance results, averaged across the tests, in Table \ref{tab:compare}. In each experiment, Flip-n-Slide improves the model performance across all evaluation metrics when compared to conventional methods. These results position Flip-n-Slide as an effective augmentation strategy for tiling large scientific images, particularly in use-cases with high class-imbalance.

\subsection{Performance Improvement for Underrepresented Classes}\label{sec:classes}
Incorporating the Flip-n-Slide strategy enhances model performance on underrepresented classes, even without tailoring the model architecture or loss function to addressing class imbalance (Table \ref{tab:compare}). Excluding the two classes in this investigation that exhibit extreme underrepresentation ($<$0.0004\% of the overall data) and are therefore subject to model noise\footnote{Flip-n-Slide also outperforms other tiling methods on the two extremely underrepresented classes, but the performance improvement is within error, so we exclude those results here.}, the next two smallest classes each make up less than 5\% of the overall dataset. Flip-n-Slide improves prediction precision by 13\%, on average, in underrepresented classes, and outperforms other approaches in every metric tested. These results highlight the improved performance for underrepresented classes, with no changes to architecture, loss function, optimization, or any other model parameters. 

Predicted masks for a withheld test set are shown for two classes in Figure \ref{fig:data_dist}. Although all tiling methods succeed at predicting the boundaries for a well-represented class (Snow \& Ice, 39.36\% of the data), only the model trained on Flip-n-Slide tiling reasonably estimates the underrepresented class (Lichen \& Moss, 4.97\%). Flip-n-Slide achieves this without altering the underlying class distribution while providing the model with more contextual views of all classes. In CNNs, classes can be distinguished by their channel spectrum or surface texture, but spatial context is also important for separating between classes \citep{wang_2023}. This is especially true in scientific use-cases where the morphometry of the scene is dictated by physical processes and can aid in class separation for rare classes. Our results confirm that the Flip-n-Slide method sufficiently generalizes a model fit to include a fuller understanding of the training distribution, especially for underrepresented classes.

\begin{table}
  \centering
  \begin{tabular}{c|cccc|ccc}
    \toprule
    \multicolumn{5}{c}{\shortstack{Performance Comparison \\ Between Strategies} } &  \multicolumn{3}{|c}{\shortstack{Performance On \\  Underrepresented Class}}\\
    \bottomrule
    \midrule
    & \multicolumn{4}{c|}{Evaluated on All Classes} & \multicolumn{3}{c}{Moss \& Lichen \textit{(4.97\% of Data)}}\\
    \midrule
    Method & IoU & Precision & F1 Score & mAP & IoU & Precision & F1 Score \\
    \midrule
    \shortstack{No Tile\\ Overlap} & 70.7\% & 71.7\% & 82.8\% & 71.3 & 42.7\% & 44.4\% & 59.9\% \\
    \midrule
    \shortstack[1]{50\% \\ Overlap} & 85.4\% & 87.1\% \% & 92.1\% & 85.5\% & 58.3\% & 62.6\% & 73.6\% \\
    \midrule
    \rowcolor{row-highlight-color} \shortstack{Flip-n-Slide \\ \textit{Ours}} & \textbf{87.6\%} & \textbf{90.1\%} & \textbf{93.4\%} & \textbf{87.8\%} & \textbf{69.7\%} & \textbf{78.4\%} & \textbf{82.2\%} \\ 
    \bottomrule 
  \end{tabular}
  \caption{Performance results for three tiling strategies across all class predictions. Flip-n-Slide outperforms other strategies in all metrics. Flip-n-Slide also improves performance results for underrepresented classes even with a basic loss function and model architecture.}
  \label{tab:compare}
\end{table}

\section{Conclusions}

In this paper, we addressed a problem with earlier tiling augmentation strategies—namely that their overlap strategy caused data redundancies, which ultimately reduce a model's ability to generalize effectively. However, tile overlapping is necessary to preserve spatial context, which is important in segmentation tasks for Earth observation use-cases. To solve this, we argued that tile overlap when combined with distinct permutations of the data may not only eliminate this redundancy but also expose the model to an expanded training dataset with more spatio-contextual views for OoI. To maintain the fidelity of this context, we emphasized the necessity of realistic augmentation choices for use in scientific segmentation tasks. We introduced a new preprocessing strategy, Flip-n-Slide, built for tiling large Earth observation images. Flip-n-Slide maintains physically realistic transformations of the input data and does not degrade the spatio-contextual information available for small OoI in the overall training set. We show that Flip-n-Slide outperforms the previous benchmarked approach at scale for tiled augmentations in all evaluation metrics, especially in cases of class imbalance, improving the detection of rare phenomena even in large imagery.

\textbf{Acknowledgements} The majority of this work was conducted on the unceded territories of the xučyun, the ancestral land of the Chochenyo speaking Muwekma Ohlone people. We have benefited, and continue to benefit, from the use of this land. We recognize the importance of taking actions in support of American Indian and Indigenous peoples who are living, flourishing members of our communities today. Testing and development of this paper was done on the CryoCloud cloud-based JupyterHub \citep{snow_2023} that is funded by the NASA Transform to Open Science Program and ICESat-2 Science Team (grant numbers 80NSSC23K0002 and 80NSSC22K1877), and on the Jupyter Meets the Earth (JMTE) cloud hub, an NSF EarthCube funded project (grant numbers 1928406 and 1928374). EA gratefully acknowledges support from a Two Sigma PhD Fellowship. We also gratefully acknowledge funding support from the NASA Cryospheric Science Program (grant number 80NSSC22K0385).

\bibliography{iclr2024_conference}
\bibliographystyle{iclr2024_conference}

\newpage
\appendix
\section{Appendix}

\subsection{The Flip-n-Slide Algorithm} \label{sec:appendix}

\begin{algorithm}
\caption{The Flip-n-Slide Algorithm.}\label{alg:cap}
\begin{algorithmic}[1]
\vspace{3px}
\Statex \textbf{Inputs:} $I$: Large Image ($x \times y$)
\Statex \quad \quad \quad \, $S$: Set of sliding tile windows \Statex \quad \quad \quad \quad \quad $(i\times j) \in (x \times y)$
\Statex \quad \quad \quad \, $F$: Set of corresponding transformations 
\Statex \quad \quad \quad \quad \quad $(i\times j) \in (x \times y)$
\vspace{2px}
\Statex \textbf{Output:} $T$: Set of augmented image tiles, $t \in T$
\vspace{3px}
\For {each ($s_{i\times j} \in S$) $\underset{i\times j}{\cap}$ ($f_{i\times j} \in F$)}:
    \State \label{alg:one} $T_t=s_{i i\times j}(I)$ 
    \State \label{alg:two} $T_t=f_{i\times j}$($s_{i\times j}$)
\EndFor
\State \textbf{return} $T$
\vspace{3px}
\end{algorithmic}
\end{algorithm}

\subsection{Computational Cost Analysis} \label{sec:cost}

In contrast to the previously accepted convention of overlap tiling, Flip-n-slide simultaneously augments and tiles the input image at the preprocessing stage, introducing a larger computational cost at this phase of the machine learning pipeline. The timing differences between the three tiling approaches used in this paper are detailed in Table \ref{tab:cost}. To find the run time, we run the $10240 \times 10240$ pixel image that we used for our segmentation task seven times through each of the three tiling strategies and take the average of these runs. We find that in both overlap approaches the $256 \times 256$ tiles are most efficient.

Although the Flip-n-Slide approach incurs a higher computational cost upfront, it streamlines downstream processes and enhances model performance during inference, particularly for underrepresented classes (Figure \ref{fig:data_dist}). By integrating data augmentation at the tiling stage of the pipeline, Flip-n-Slide also eliminates the need for additional augmentation at train time, thereby reducing computational costs at the training stage. Furthermore, many approaches that employ the previous overlap convention at training time recommend using the same overlapping tiling approach for prediction averaging at each pixel during test time \cite{unel_2019, zeng_2019, reina_2020, akyon_2022}. However, this necessitates tracking pixel locations from tile to tile and complicates the process of reconstructing test tiles back into the overall input image, incurring further computational overhead at the inference stage of the pipeline. Therefore, despite the increased computational cost associated with the Flip-n-Slide strategy during preprocessing, its performance improvements for underrepresented classes justify the upstream computational investment. Depending on the choices that a user makes throughout the rest of the pipeline, both at train and test time, these initial costs may result in reduced overhead later on.

\begin{table}[b]
  \centering
  \begin{tabular}{ccccc}
    \toprule
    \toprule
    \multicolumn{5}{c}{Tiling Runtime for $10240 \times 10240$ Input Image} \\
    \midrule
    Method & $64 \times 64$ Tiles & $128 \times 128$ Tiles & $256 \times 256$ Tiles & $512 \times 512$ Tiles \\
    \midrule
    No Overlap & 4.9s & 4.3s & 4.2s & 4.0s\\
    \midrule
    50\% Overlap & 16.4s & 14.9s & 14.9s & 16.3s \\
    \midrule
     Flip-n-Slide & 71s & 63s & 59.3s & 61s \\ 
    \bottomrule 
  \end{tabular}
  \caption{Computational cost for simultaneously tiling and augmenting a large input image using the Flip-n-Slide approach. Although Flip-n-Slide has an increased upstream cost, its performance improvements for underrepresented classes justify the upstream computational investment and these initial costs may result in reduced overhead later on.}
  \label{tab:cost}
\end{table}

\subsection{Ablation Studies} \label{sec:ablation}

\begin{figure*}
  \centering
  \includegraphics[width=0.65\linewidth]{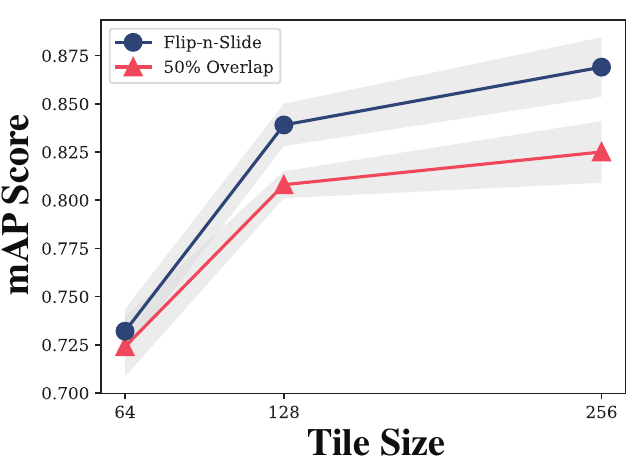}
  \caption{Ablation studies on varying tile size confirm that our strategy leads to expected model behavior. Here we show mAP for each study, averaged over three test runs with the standard deviation shown in grey. Flip-n-Slide outperforms the conventional 50\% overlap strategy even at smaller input tile sizes.}
  \label{fig:ablation}
\end{figure*}

Ablation studies are necessary to disambiguate between the role of various fixed decisions on the model outcomes. We explore the impacts of varying tile size choice on the model performance. Previous studies show that larger tiles are more effective for segmentation tasks under traditional augmentation strategies \cite{zeng_2019,reina_2020}. We test the performance of the Flip-n-Slide strategy across a range of two additional tile sizes: $64 \times 64$, $128 \times 128$. Due to GPU size limitations we could not test tiles of $512 \times 512$ at the same batch size, so we do not include it here. We reproduce the results of previous studies, finding that larger tile size does lead to better performance. However, we find that even at smaller tile sizes, our approach outperform previous strategies even when they are implemented at larger sizes. When $128 \times 128$ tile sizes are generated with Flip-n-Slide they perform at a better score to $256 \times 256$ tiles generate by a 50\% overlap strategy (Figure \ref{fig:ablation}). 

\begin{figure}
  \centering
  \includegraphics[width=0.45\linewidth]{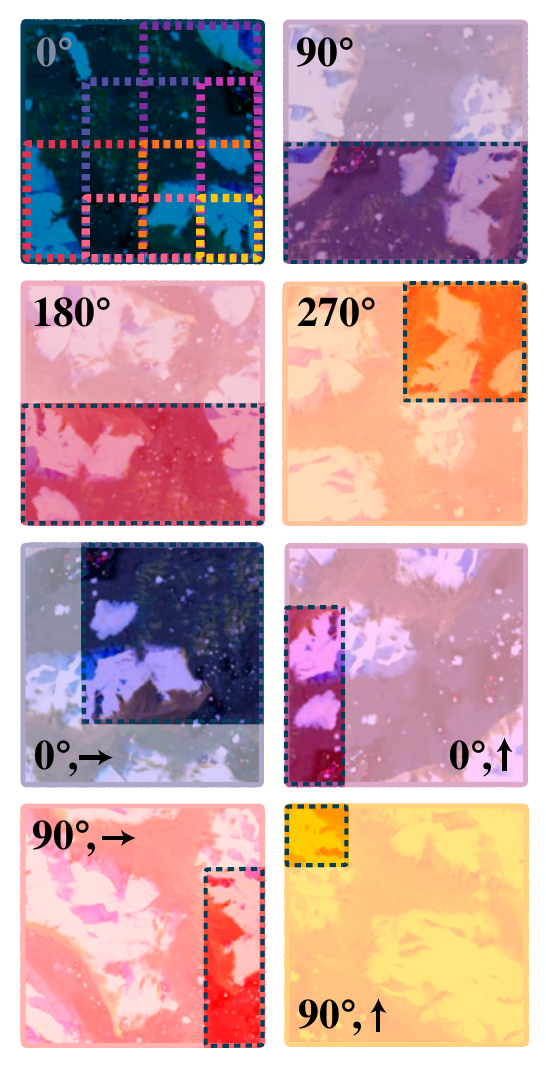}
  \caption{To minimize redundancy in the Flip-n-Slide strategy, each overlapping tile is uniquely permuted with a distinct, physically-realistic transformation, as shown here. Previous strategies have not employed overlap-specific transformations; any augmentations have been applied across all tiles or at random. Tiles are shown in false color to illustrate overlapping areas. Transparency indicates areas that do not overlap with the blue tile shown here. They overlap with neighboring blue tiles instead.}
  \label{fig:permute}
\end{figure}

\end{document}